\documentclass[10pt,twocolumn,letterpaper]{article}

\usepackage{cvpr}              

\usepackage{graphicx}
\usepackage{amsmath}
\usepackage{amssymb}
\usepackage{booktabs}
\usepackage{arydshln}
\usepackage{times}
\usepackage{epsfig}
\usepackage{graphicx}
\usepackage{color}
\usepackage{xcolor}
\usepackage{framed}
\usepackage{color, colortbl}
\usepackage{setspace}
\usepackage{multirow}
\usepackage{pifont}

\usepackage{enumitem}
\usepackage{bbm}
\usepackage{dsfont}

\usepackage{algorithm}
\usepackage{algpseudocode}

\usepackage[pagebackref,breaklinks,colorlinks]{hyperref}

\colorlet{shadecolor}{yellow!20}

\usepackage[capitalize]{cleveref}
\crefname{section}{Sec.}{Secs.}
\Crefname{section}{Section}{Sections}
\Crefname{table}{Table}{Tables}
\crefname{table}{Tab.}{Tabs.}

\def\cvprPaperID{25} 
\def\confName{CVPR}
\def\confYear{2022}

\begin{document}


\title{Lost in Compression: the Impact of Lossy Image Compression \\ on Variable Size Object Detection within Infrared Imagery}

\author{Neelanjan Bhowmik\textsuperscript{1}, Jack W. Barker\textsuperscript{1},  Yona Falinie A. Gaus\textsuperscript{1}, Toby P. Breckon\textsuperscript{1,2}\\
Department of \{Computer Science\textsuperscript{1}, Engineering\textsuperscript{2}\}, Durham University, Durham, UK \\
}

\maketitle

\begin{abstract}
\noindent
Lossy image compression strategies allow for more efficient storage and transmission of data by encoding data to a reduced form. This is essential enable training with larger datasets on less storage-equipped environments. However, such compression can cause severe decline in performance of deep Convolution Neural Network (CNN) architectures even when mild compression is applied and the resulting compressed imagery is visually identical. In this work, we apply the lossy JPEG compression method with six discrete levels of increasing compression \{95, 75, 50, 15, 10, 5\} to infrared band (thermal) imagery. Our study quantitatively evaluates the affect that increasing levels of lossy compression has upon the performance of characteristically diverse object detection architectures (Cascade-RCNN, FSAF and Deformable DETR) with respect to varying sizes of objects present in the dataset. When training and evaluating on uncompressed data as a baseline, we achieve maximal mean Average Precision (mAP) of $0.823$ with Cascade R-CNN across the FLIR dataset, outperforming prior work. 
The impact of the lossy compression is more extreme at higher compression levels ($15, 10, 5$) across all three CNN architectures. However, re-training models on lossy compressed imagery notably ameliorated performances for all three CNN models with an average increment of $\sim76\%$ (at higher compression level $5$). Additionally, we demonstrate the relative sensitivity of differing object areas \{tiny, small, medium, large\} with respect to the compression level. We show that tiny and small objects are more sensitive to compression than medium and large objects. Overall, Cascade R-CNN attains the maximal mAP across most of the object area categories.

\end{abstract}


\vspace{-0.6cm}
\section{Introduction} \label{sec:intro}
\vspace{-0.2cm}

\begin{figure}[!htb]
    \centering
    \includegraphics[width=\linewidth]{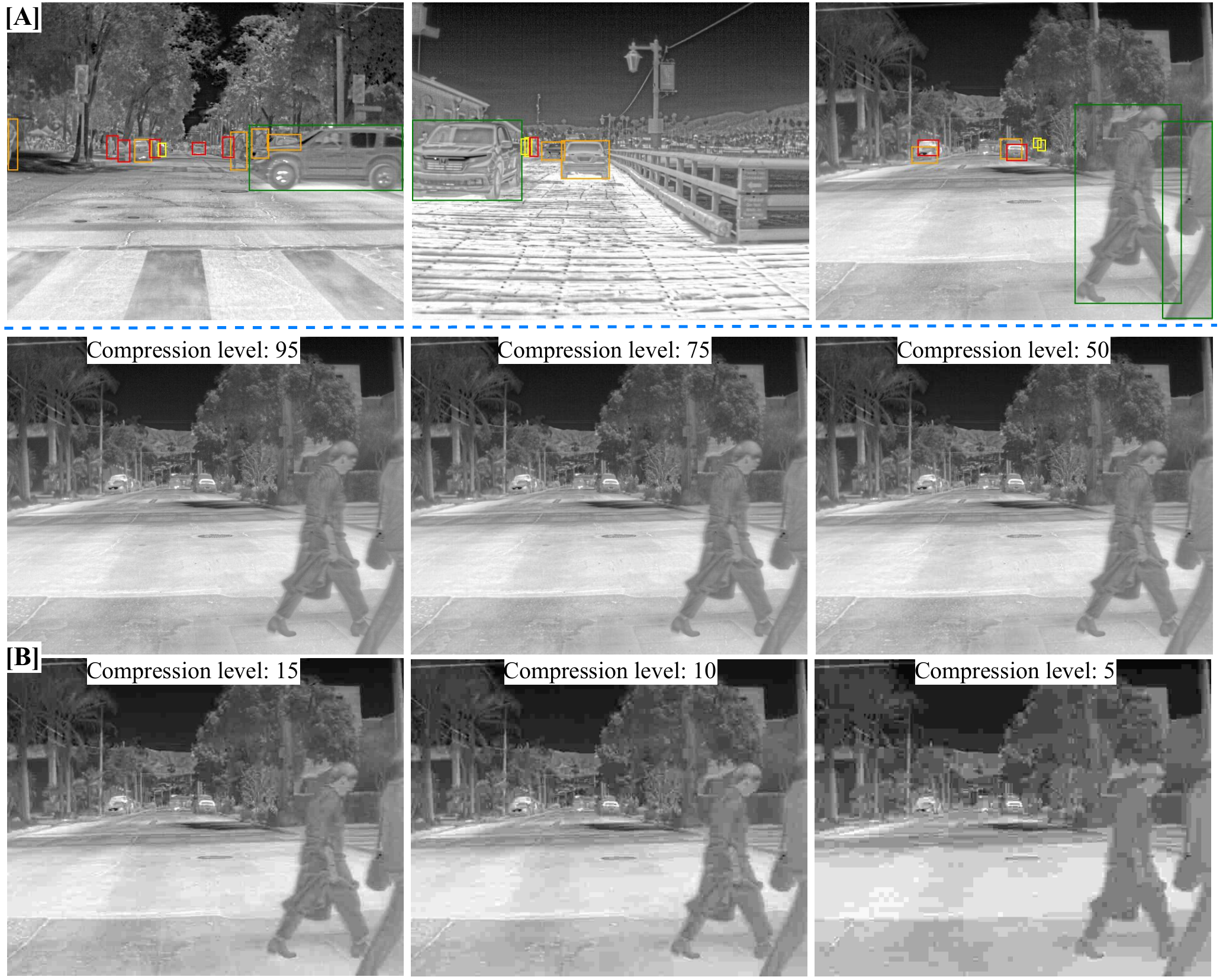}
    \vspace{-0.8cm}
    \caption{Exemplar infrared imagery from FLIR \cite{flir2019flir} containing different classes  with varying object areas in bounding boxes: \textcolor{yellow}{{\it tiny}}, \textcolor{red}{{\it small}}, \textcolor{orange}{{\it medium}}, \textcolor{green}{{\it large}} [A]. Example imagery at different lossy compression levels [B].}
    \vspace{-0.4cm}
    \label{fig:flir_ex}
\end{figure}
\noindent
The use of infrared-band (thermal) camera imagery (Figure \ref{fig:flir_ex}) within the task of visual surveillance is well established with applications within target detection, visual tracking, behaviour analytics, home monitoring, and automotive environment perception \cite{smeulders2013visual,kundegorski2014photogrammetric,kundegorski2016real,kundegorski2015posture, li2010robust,teutsch2014low, brehar2014pedestrian}.
 The synergistic nature of combining both visible-band and thermal-band imagery is advantageous when 
applied to a range of computer vision applications \cite{lin2001extending, loveday2018impact}. Valuable visual
cues can be provided in the infrared domain by thermal
sensors which visible-band sensors will fail to capture\cite{mukhtar2015vehicle, hwang2015multispectral}.
A major advantage of thermal imagery is that it is rarely influenced by surrounding
lighting changes and shadows, meaning that objects-of-interest can be readily distinguished in the dark, fog and other complex environments as opposed
to the scene illumination requirements of visible-band imagery.

The recent rise of Convolutional Neural Networks (CNN) \cite{krizhevsky2012imagenet} have revolutionised visual tasks significantly advancing the state-of-the-art in many applications.
Within object detection specifically,
most efforts have focused on detecting objects-of-interest in standard colour imagery  by  using multi-stage \cite{ren17:fasterrcnn,MaskRCNNHe2017, cai19:cascade}, singe-stage \cite{redmon2018yolov3, lin18:retinanet, zhu19:fsaf} and transformer-based \cite{carion20:detr, zhu2021:deformable} detectors.
These aforementioned object detection based CNN methods rely heavily on architectures that have been trained on large-scale colour imagery datasets such as ImageNet \cite{deng2009imagenet}
PASCAL Visual Object Classes (PASCAL-VOC) \cite{everingham2010pascal} or Microsoft Common Objects in Context (MS-COCO) \cite{lin15:coco}. 
Introducing CNN to object detection within thermal imagery is significantly hindered by the absence of such annotated datasets of the same scale and variety. In order to combat this challenge, methods such as transfer learning \cite{gaus2020visible}, generation of pseudo-RGB equivalents \cite{devaguptapu2019borrow}, and domain adaptation \cite{munir2021sstn} are used to benefit an existing CNN model and raise an equivalent level of CNN success for thermal band imagery. The work of \cite{gaus2020visible} focuses on thermal object  detection  by employing a transfer learning approach. In this approach, the  knowledge  obtained from the  visible  spectrum  is transferred to  the thermal  domain  for object  detection  in  the  thermal  domain. The work of \cite{devaguptapu2019borrow} implements image-to-image translation frameworks to generate pseudo-RGB equivalents of a given thermal image and then use a CNN architecture for object detection in the thermal image. Recent work of \cite{munir2021sstn} proposes  a  self-supervised  domain  adaptation via an encoder-decoder  transformer network  to  develop  a  robust  thermal  image  object detector in autonomous driving.

Most of the solutions stated above have made a step towards real-time processing which is vital to the application of autonomous driving. Autonomous vehicles and other such agents can utilise thermal-band sensors to provide a necessary solution to accurately  perceive the  environment even in differing light conditions, or night time. However, the real-time transmission of high-resolution visible-band and thermal-band data collected by the agent during inference can require substantial transmission overhead to process in real-time, and large storage requirements for storing such data. Every fraction of a second of reaction time counts for autonomous vehicles dealing with unexpected obstructions in the path of motion, so reducing the transmission overhead is vital for improved safety. The use of commonplace lossy compression techniques, such as JPEG \cite{wallace1992jpeg} and  MPEG \cite{le1991mpeg} tackles such transmission and storage overheads by compressing the data to a reduced form; a downside of which is potential image quality degradation (Figure \ref{fig:flir_ex}, B) which can affect the performance of computer vision models. Work by \cite{dodge2016understanding} evaluates the affect of five types  of  quality  distortions and compression (blur, noise, contrast,  JPEG and JPEG2000) using  CNN  architectures  when  classifying  a subset of the ImageNet dataset \cite{deng2009imagenet}. More recent work by \cite{poyser20:lossy} thoroughly evaluates the  impact  of  JPEG  and  H.264  lossy  compression on CNN  architectures. They  evaluate  Faster  R-CNN  \cite{ren17:fasterrcnn}  upon the  Pascal  VOC  dataset \cite{everingham2010pascal} and find similarly  to  \cite{dodge2016understanding}  that performance degrades rapidly at high lossy compression levels.

The prior work on image compression stated above largely focuses on the use of compression applied to visible-domain imagery. Relatively few studies  investigate  the  impact  upon  CNN  task performance  with  respect  to  differing  levels  of  compression applied to infrared-band (thermal) at inference (deployment) time. In  this  work  we  make  the following contributions:
\vspace{-0.2cm}
\begin{itemize}
  \item[--] we  investigate  three  diverse end-to-end  CNN  object  detection architectures,  which  differ  in  operation,  i.e.,  multi-stage (Cascade  R-CNN  \cite{cai19:cascade}),  single-stage  (FSAF \cite{zhu19:fsaf}) and transformer-based (Deformable DETR \cite{zhu2021:deformable}) over thermal imagery.
  \vspace{-0.3cm}
  \item[--] as thermal imagery exhibits inherently  different  properties  from  standard  visible-band imagery,  we  examine  the  impact that lossy  image compression at differing levels has upon  CNN-based thermal object detection.
  \vspace{-0.3cm}
  \item[--] furthermore, we thoroughly examine  the impact of lossy image compression with respect to in-image object size definition, (\textit{tiny, small, medium, large}) and determine within which domains compression is most impactful upon performance and hence where image quality is most pertinent to deployed object detection model performance.

\end{itemize}

\vspace{-0.4cm}
\section{Related Work} \label{sec:soa}
\vspace{-0.2cm}
\noindent
Object detection and classification in thermal imagery is an active area of research \cite{peng2016nirfacenet, lee2016recognizing, rodger2016classifying}. A range of trial works in the literature address the task of detecting people and objects in thermal imagery \cite{hwang2015multispectral, gaus2020visible, devaguptapu2019borrow}. There has been a significant
amount of work on classifying and detecting people and
objects in thermal imagery using standard computer vision
and machine learning models. An early method such as a template-based approach, where
Bertozzi et al, \cite{bertozzi2007pedestrian} implements probabilistic human shape
templates whilst Davis and Keck \cite{davis2005two} uses generalized
person templates derived from contour saliency maps for pedestrian detection under thermal imagery. In \cite{li2010robust}, features decomposed by the wavelet transform from  the high brightness property of the pedestrian pixels under thermal imagery were used as an input to a support vector machine (SVM) classifier.

With the increasing popularity of deep CNN architectures, several methods have been proposed for applying deep learning methods to thermal imagery \cite{gaus2020visible, devaguptapu2019borrow, peng2016nirfacenet, lee2016recognizing, rodger2016classifying}. In most cases, this research was carried out in the field of autonomous driving, where accurate detection of pedestrians and vehicles is vital. A transfer learning approach with the YOLO \cite{redmon16yolo9000} architecture has been carried out by Abbot et al. \cite{abbott2017deep}, in which high-resolution thermal imagery is used for training and low-resolution thermal is used for evaluation purpose (to classify pedestrians and vehicles).
Devaguptapu et al.\cite{devaguptapu2019borrow} address the data scarcity problem in thermal imagery by utilising
image-to-image translation frameworks \cite{zhu2017unpaired} to generate pseudo-RGB equivalents of given thermal imagery, then
employing Faster-RCNN \cite{ren17:fasterrcnn} for detecting object-of-interest.
Faster-RCNN \cite{ren17:fasterrcnn} has also been trained with  thermal imagery under  a
super-resolution method to deal with the issue of a small number of pixels which targets at the long-range have \cite{zhang2018novel}. Chao et al. \cite{cao2019every} take another approach by proposing the one-stage detector
ThermalDet which utilises all the features in different levels of the feature pyramid extracted by the
backbone network, resulting in higher detection accuracy than baseline (Faster R-CNN)  \cite{ren17:fasterrcnn}.

Some works propose
the use of thermal imagery as a complement of colour imagery by fusing both domains, achieving superior results.
The work of Liu et al. \cite{liu2016multispectral} integrates features from  both colour and thermal at different stages, resulting in better detection accuracy when compared to the baseline (Faster R-CNN). A fusion method was also explored via  multi-layer fusion RPN. This was used for integrating features
in different branches, significantly  reducing  the  detection  miss  rate in several challenging datasets \cite{chen2018multi}. Later work such as \cite{sun2019rtfnet} fuses both visible and thermal imagery via a fusion-based network for the semantic segmentation of urban scenes. Namely the results from RGB-Thermal Fusion Network (RTFNet) demonstrate the superiority of the such an approach,
even in challenging lighting conditions. Similar to RTFNet, Multi-spectral Fusion Networks (MFNet) architecture \cite{ha2017mfnet} which fuses visible and thermal imagery resulting in similar or higher accuracy than
state-of-the-art segmentation methods such as SegNet.

While these efforts have shown good detection accuracy performance under thermal imagery, implementing deep CNN architectures requires heavy memory usage, substantial storage
and transmission infrastructure. There has
been substantial investigation of efficient storage and reducing memory utilisation \cite{poyser2021impact, webb2021operationalizing, dodge2016understanding} but very few works have examined the performance of these techniques
applied to thermal imagery. These  limited  studies  open  the  door  only  slightly  on  the  question - \textit{what is the generalised impact of lossy compression across a diverse set of deep neural network object detection architectures with respect to varying object size within infrared-band (thermal) imagery?}

\vspace{-0.3cm}
\section{Proposed Approach}  \label{sec:proposal}
\vspace{-0.2cm}
\noindent
We outline the approach of this paper in the following sections. Our method utilises the JPEG compression method (Section \ref{ss:compression}) across infrared-band (thermal) image data. We iterate through six discrete levels of compression and compare the effect that each level has on the performance of object detection architectures (Section \ref{ss:objarch}).   

\subsection{Object Detection Architectures} \label{ss:objarch}
\vspace{-0.2cm}
\noindent
In this study we utilise three state-of-the-art, well-established and characteristically diverse object detection architectures as outlined in Table \ref{Tab:objdet}. These are namely: Cascade R-CNN \cite{cai19:cascade}, Feature Selective Anchor-Free \cite{zhu19:fsaf}, and Deformable End-to-End Detection with Transformers \cite{zhu2021:deformable}. All architectures in this study utilise a ResNet-50 \cite{He15:ResNet} backbone and are initialised with weights trained on COCO \cite{lin15:coco}.

\begin{table}[!htb]
	\centering
	\renewcommand*{\arraystretch}{0.90}
	\caption{Summary of object detection architectures.}
	\vspace{-0.4cm}
	\begin{tabular}{|l|l|}
		\hline 
		Architecture & Key Features \\ \hline
		
		Cascade R-CNN \cite{cai19:cascade} & two-stage, anchor-based \\ 
		FSAF \cite{zhu19:fsaf} & single-stage, anchor-free \\ 
		\multirow{2}{*}{Deformable DETR \cite{zhu2021:deformable}} & transformer-based  \\ 
		& single-stage, anchor-free \\ \hline
		
		
	\end{tabular}
	\label{Tab:objdet}  
\end{table}

\noindent {\bf Cascade R-CNN (CR-CNN) \cite{cai19:cascade}:} Cascade Region-based Convolutional Neural Network (Cascade R-CNN) is an extension of R-CNN \cite{RCNN-Girshick2014} which offers a solution to the trade-off between low Intersection over Union (IoU) thresholds inducing noisy detections and performance degradation with high IoU thresholds. It accomplishes this by training a sequence of detectors stage-by-stage with increasing IoU thresholds to be more selective against false positives.

\noindent {\bf FSAF \cite{zhu19:fsaf}:} Feature Selective Anchor-Free (FSAF) is a module for single-shot object detection can be added to detectors with a feature pyramid structure and performs online feature selection upon multi-level anchor-free branches. This aims to address the limitations induced by heuristic-guided feature selection and overlap-based anchor sampling, leading to increased performance while introducing negligible inference overhead.

\noindent {\bf Deformable DETR (DDETR) \cite{zhu2021:deformable}:} Detection Transformer (DETR) combines convolutional features with a transformer architecture \cite{Vaswani2017Transformer} which powerfully model sequential relations using multi-head attention. This means that the DETR architecture does not rely on hand-crafted components and as such can be trained fully end-to-end. Deformable DETR is an extension to this architecture which speeds up the convergence by having attention modules only attend to a small set of neighbouring points as well as tackling the problem of representing objects at varying scale.

\subsection{Lossy Image Compression} \label{ss:compression}
\vspace{-0.2cm}
\noindent
For this work we use the lossy JPEG \cite{wal91:jpeg} compression algorithm which is based on the discrete cosine transform (DCT). The first step of JPEG is colour conversion where the RGB data of the image is converted using the respective components of luminance (Y), blue projection (U) and red projection (V). The resulting YUV image is then split into $8 \times 8$ pixel blocks to be processed into the frequency domain by the Discrete Cosine Transform (DCT). The next step of quantisation is where the reduction of information required to store the image takes place in which the resultant DCT coefficient matrix is divided and rounded by the quantisation matrix leading to a reduced form that provides resolution amount with respect to how perceivable a given image part is. 

The quantisation is where the image quality of the resulting compressed representation can be controlled by the compression level parameter. If the value of this parameter is $1$, maximum compression will produce the lowest quality of the data but will yield the smallest file size required to store the data. Conversely, a value of $100$ offers the least effective compression which will negligibly affect the quality of the resultant image, producing a visually lossless image and as such the file size will remain close to the original image prior to compression.

\subsection{Object Area Definition} \label{ss:objarea}
\vspace{-0.2cm}
\noindent
The object detection dataset comprises of several object classes with varying bounding box area size. In this work, we further analyse each object class based on the object area size. We adhere to the COCO \cite{lin15:coco} benchmark definitions of three discrete categories of the object areas: {\it small, medium, large}. Additionally, we introduce a new category, {\it tiny}, where the object area is less than or equal to $20^2$ pixels. The object areas (Figure \ref{fig:flir_ex}, A) used in this work are outlined in Table \ref{Tab:objarea}. In this work, we incorporate the thorough study of the impact of the lossy image compression on the four defined object areas (Table \ref{Tab:objarea}), and outline the detection performance of CNN architectures (Section \ref{ss:objarch}).
\vspace{-0.6cm}
\begin{table}[!htb]
\centering
\renewcommand*{\arraystretch}{0.90}
\caption{Object area definition.}
\vspace{-0.4cm}
\begin{tabular}{|l|l|}
\hline 
Object & Area \\ \hline
{\it tiny} & area $\leq 20^2$ \\ 
{\it small} & area $\leq 32^2$ \\ 
{\it medium} & $32^2 <$ area $\leq 96^2$ \\ 
{\it large} &  $96^2 < $ area \\ \hline

\end{tabular}
\label{Tab:objarea}  
\end{table}


\vspace{-0.5cm}
\section{Experimental Setup} \label{sec:eva}
\vspace{-0.2cm}
\noindent
This section presents the dataset used, experimental strategy and the implementation details of our experiments.

\subsection{Dataset} \label{ss:dataset}
\vspace{-0.2cm}
\noindent
The experimental setup comprises of following dataset.
\noindent {\bf FLIR \cite{flir2019flir}}: The \textit{FLIR} dataset provides annotated single channel grayscale infrared imagery (Figure \ref{fig:flir_ex}) of multiple object classes, which are captured under clear-sky conditions during both day (60\%) and night (40\%). Thermal images are acquired with a FLIR Tau2 camera (Long Wave Infrared Cameras - LWIR) with image resolution of $(640 \times 512)$. In this work, we use the default training and testing split provided in the dataset. We consider primarily three classes, \{{\it Person, Bicycle, Car}\}, for our object detection task. The training and testing set consist of $7,859$ and $1,360$ images respectively. The details of the FLIR dataset statistics are presented in Table \ref{Tab:flirthermal_dbstat}. The object area-wise (Section \ref{ss:objarea}) statistics of each class is illustrated in Table \ref{Tab:flirthermal_dbstat}. 

\begin{table}[!htb]
\centering
\renewcommand*{\arraystretch}{0.90}
\caption{FLIR \cite{flir2019flir} dataset statistics: class-wise number of instances, \textcolor{blue}{\%} of \{Train, Test\} set.}
\vspace{-0.4cm}
\addtolength{\tabcolsep}{-1.0pt}
\begin{tabular}{|c|c|c|c|}
\cline{2-4} 
\multicolumn{1}{l|}{} & \small Person & \small Bicycle & \small Car \\ \hline 


\multicolumn{1}{|l|}{\small Train-set} & 
\small 22,372 &
\small 3,986 &
\small 41,260 \\ \hline 
\multicolumn{1}{|l|}{\small Obj$_{tiny}$} &
\small 8923, \textcolor{blue}{13.15\%} & 
\small 993, \textcolor{blue}{1.46\%} & 
\small 10832, \textcolor{blue}{15.97\%} \\ 
\multicolumn{1}{|l|}{\small Obj$_{small}$} &
\small 16304, \textcolor{blue}{24.03\%} & 
\small 2419, \textcolor{blue}{3.57\%} & 
\small 20665, \textcolor{blue}{30.46\%} \\ 
\multicolumn{1}{|l|}{\small Obj$_{med}$} &
\small 5590, \textcolor{blue}{8.24\%} & 
\small 1505, \textcolor{blue}{2.22\%} & 
\small 17056, \textcolor{blue}{25.14\%} \\ 
\multicolumn{1}{|l|}{\small Obj$_{large}$} &
\small 478, \textcolor{blue}{0.7\%} & 
\small 62, \textcolor{blue}{0.09\%} & 
\small 3539, \textcolor{blue}{5.22\%} \\ \hline \hline

\multicolumn{1}{|l|}{\small Test-set} & 
\small 5,779 &
\small 471 &
\small 5,432 \\ \hline

\multicolumn{1}{|l|}{\small Obj$_{tiny}$} &
\small 765, \textcolor{blue}{6.54\%} & 
\small 55, \textcolor{blue}{0.47\%} & 
\small 807, \textcolor{blue}{6.9\%} \\ 
\multicolumn{1}{|l|}{\small Obj$_{small}$} &
\small 2885, \textcolor{blue}{24.67\%} & 
\small 219, \textcolor{blue}{1.87\%} & 
\small 2248, \textcolor{blue}{19.22\%} \\ 
\multicolumn{1}{|l|}{\small Obj$_{med}$} &
\small 2575, \textcolor{blue}{22.02\%} & 
\small 230, \textcolor{blue}{1.97\%} & 
\small 2465, \textcolor{blue}{21.08\%} \\ 
\multicolumn{1}{|l|}{\small Obj$_{large}$} &
\small 319, \textcolor{blue}{2.73\%} & 
\small 22, \textcolor{blue}{0.19\%} & 
\small 719, \textcolor{blue}{6.15\%} \\ \hline \hline

\multirow{2}{*}{Images} & \multicolumn{2}{c|}{Train} & \multicolumn{1}{c|}{Test} \\ \cline{2-4}
& \multicolumn{2}{c|}{7,859} & \multicolumn{1}{c|}{1,360} \\ \hline

\end{tabular}
\addtolength{\tabcolsep}{-0pt} 
\label{Tab:flirthermal_dbstat}  
\end{table}

\noindent \textbf{Compressed datasets.} To determine how much lossy JPEG compression \cite{wal91:jpeg} is achievable within CNN object detection architectures on infrared imagery, the original uncompressed \textit{FLIR} \cite{flir2019flir} dataset is compressed at six different levels \{{\it 95, 75, 50, 15, 10, 5}\} to create the compressed versions of the dataset ({\it FLIR$\_c$}), as depicted in (Figure \ref{fig:flir_ex}, B). 

\subsection{Experimental Protocol} \label{ss:expprotocol}
\vspace{-0.2cm}
\noindent
Our experiments consist of the following settings:
\vspace{-0.2cm}
\begin{itemize}
    \item [--] Firstly, the three CNN architectures (Section \ref{ss:objarch}) are trained and evaluated on uncompressed imagery, ({\it FLIR} $\Rightarrow$ {\it FLIR}), for benchmark purpose. 
    \vspace{-0.3cm}
    \item [--] Secondly, the CNN architectures are trained on uncompressed imagery and evaluated on the compressed imagery, ({\it FLIR} $\Rightarrow$ {\it FLIR$\_c$}).
    \vspace{-0.3cm}
    \item [--] Finally, each CNN architecture is re-trained with compressed imagery at each of the six lossy compression levels to determine whether resilience to compression could be improved, and how much compression can be achieved before a significant impact on object detection performance is observed ({\it FLIR$\_c$} $\Rightarrow$ {\it FLIR$\_c$}).
\end{itemize}

\subsection{Implementation Details} \label{ss:implementation}
\vspace{-0.0cm}
\noindent
The CNN architectures (Section \ref{ss:objarch}) are implemented using the MMDetection framework \cite{mmdetection}. All experiments are initialised with weights pretrained on the COCO dataset \cite{lin15:coco}. The CNN architectures (Section \ref{ss:objarch}) are trained using a ResNet$_{50}$ \cite{He15:ResNet} backbone with the following training configuration: 
\vspace{-0.2cm}
\begin{itemize}
    \item[--] Cascade R-CNN \cite{cai19:cascade} and FSAF \cite{zhu19:fsaf}: backpropagation optimisation is performed via Stochastic Gradient Descent (SGD), with initial learning rates of ${1e-2}$, trained for $20$ epochs.
    \vspace{-0.2cm}
    \item[--] Deformable DETR \cite{zhu2021:deformable}: backpropagation optimisation is performed via the Adam optimiser, with initial learning rates of ${1e-4}$, trained for $50$ epochs.
\end{itemize}

Standard data augmentation techniques, such as Random Crop, Random Flip, have applied during model training with an application probability of $0.5$.

\vspace{-0.2cm}
\section{Results} \label{sec:result}
\vspace{-0.0cm}
\noindent
The model performance is evaluated through MS-COCO metrics \cite{lin15:coco}, with IoU greater than $0.5$, using Average Precision (AP) for class-wise, and mAP for the overall performance measurement. Additionally, we compare model performance via: Complexity (number of parameters in millions, C), the ratio between mAP and the number of parameters in the architecture (mAP:C), and the inference time in milliseconds (ms) taken by the respective model to process an individual frame. Model inference is carried out on the NVIDIA 1080Ti GPU. The highlighted values in each table denote the maximal performance achieved.

\subsection{Benchmarking on Infrared dataset} \label{ss:baselineflir}
\vspace{-0.2cm}
\begin{table}[!htb]
\centering
\renewcommand*{\arraystretch}{0.90}
\vspace{-0.3cm}
\caption{FLIR: Performance (class name indicates AP, mAP indicates mean average precision of all classes) of CNN models.}
\vspace{-0.4cm}
\addtolength{\tabcolsep}{-3.4pt}
\begin{tabular}{|l|l|l|l|l|}
\hline
\multirow{2}{*}{Model} & \multicolumn{3}{c|}{Average Precision (AP)} & \multirow{2}{*}{mAP} \\ \cline{2-4}
&  Person & Bicycle & Car & \\  \hline 
MMTOD-UNIT \cite{Devaguptapu_2019:borrow} & 0.644 & 0.494 & 0.707 & 0.615  \\
ThermalDet \cite{yu19:thermaldet}  & 0.782 & 0.600 & 0.855 & 0.746  \\
Pseudo-two-stage \cite{ZHOU2021:flir} & 0.787 & 0.624 & 0.855 & 0.755 \\ \cdashline{1-5}
CR-CNN & {\bf 0.877} & {\bf 0.681} & 0.911	& {\bf 0.823}   \\ 
FSAF & 0.867	& 0.675	& 0.904	& 0.815 \\
DDETR & 0.863 & 0.659 & {\bf 0.913} & 0.812 \\ \hline

\end{tabular}
\addtolength{\tabcolsep}{-0pt} 
\label{Table:mAP_flir}
\end{table}



Table \ref{Table:mAP_flir} presents object detection performance in infrared imagery for the first set of experiments using the CNN architectures set out in Section \ref{ss:objarch}. These models are applied to the \textit{FLIR} \cite{flir2019flir} dataset to provide benchmark performance. The best performance on {\it FLIR} (mAP: $0.823$, Table \ref{Table:mAP_flir}, lower) is obtained by Cascade R-CNN \cite{cai19:cascade} achieving the highest AP on two classes ({\it Person, Bicycle}) out of three. FSAS \cite{zhu19:fsaf} and DDETR \cite{zhu2021:deformable}, also produce comparable mAP ($0.815$ and $0.812$), where DDETR achieves the maximal AP on the {\it Car} class. All three models significantly outperform the prior works of \cite{Devaguptapu_2019:borrow, yu19:thermaldet, ZHOU2021:flir} (mAP: $0.615, 0.746, 0.755$, Table \ref{Table:mAP_flir}, upper).
Overall, CR-CNN marginally outperforms the one-stage and transformer-based models in this work. This is possibly due to the architectural design of CR-CNN, where images are sampled with increasing IoU thresholds to tackle different training distributions.

\begin{table}[htb!]
\renewcommand*{\arraystretch}{0.90}
    \centering
    \caption{FLIR: Model complexity and efficiency.}
    \vspace{-0.4cm}
    \begin{tabular}{|l|l||l|l|l|}
    \hline
    Model & C & mAP & mAP:C  & ms \\ \hline
    
    CR-CNN & 69.16 & 0.823 & 0.011  & 93.20 \\ 
    FSAF & 36.24 & 0.815 & {\bf 0.022}  & {\bf 64.69} \\
    DDETR & 40.09 & 0.812 & 0.020 & 104.98  \\ \hline \hline
    
    \multicolumn{5}{|l|}{Inference on: NVIDIA $1080Ti$ GPU} \\ \hline
    
    \end{tabular}
    \label{tab:flircomplexity}
\end{table}

Additionally, we present the computational efficiency, and speed, which are crucial criteria for operational and real-world deployment perspectives in Table \ref{tab:flircomplexity}. Being the smallest model, FSAF \cite{zhu19:fsaf}, which has $\sim2\times$ fewer parameters than Cascade R-CNN \cite{cai19:cascade}, obtains the maximal computational efficiency (mAP:C) of $0.022$ (Table \ref{tab:flircomplexity}). All three models achieve real-time throughput with FSAF obtaining the fastest individual frame processing speed of 64.69ms, $\sim1.5\times$ faster compared to Cascade R-CNN \cite{cai19:cascade} and Deformable DETR \cite{zhu2021:deformable} which obtain 93.2ms and 104.98ms respectively across the same metric.


\subsection{Impact of Lossy Compression} \label{ss:lossy}
\vspace{-0.1cm}
\begin{figure*}[!htb]
    \centering
    \includegraphics[width=\linewidth]{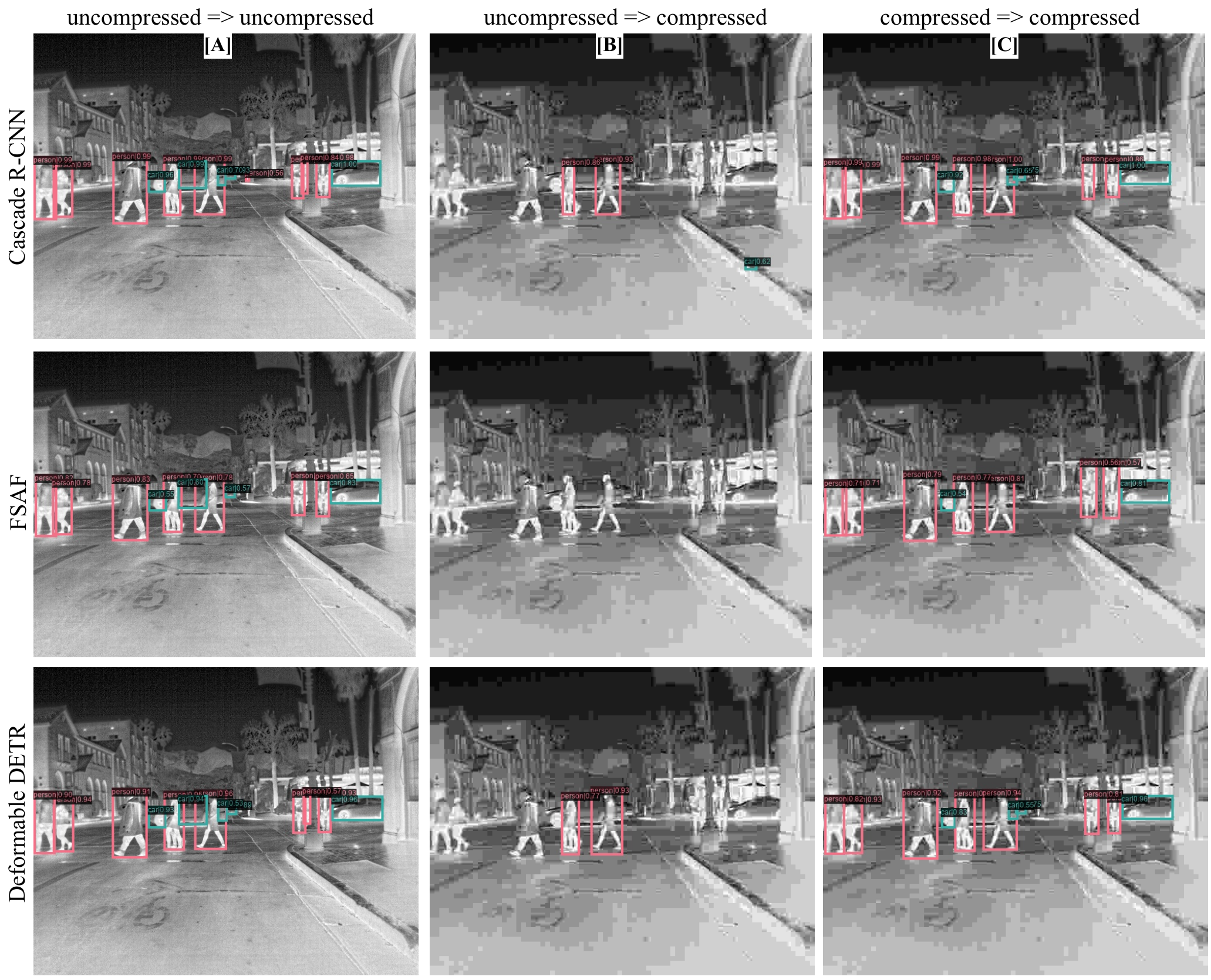}
    \vspace{-0.8cm}
    \caption{Detection using varying CNN models: [A] uncompressed settings ({\it FLIR $\Rightarrow$ FLIR}), [B] {\it FLIR $\Rightarrow$ FLIR$\_c$} at compression level $5$, [C] re-training with compressed images at compression level $5$ ({\it FLIR$\_c$ $\Rightarrow$ FLIR$\_c$}).}
    \label{fig:flirdet_ex}
\end{figure*}

\begin{figure*}[!htb]
    \centering
    \includegraphics[width=\linewidth]{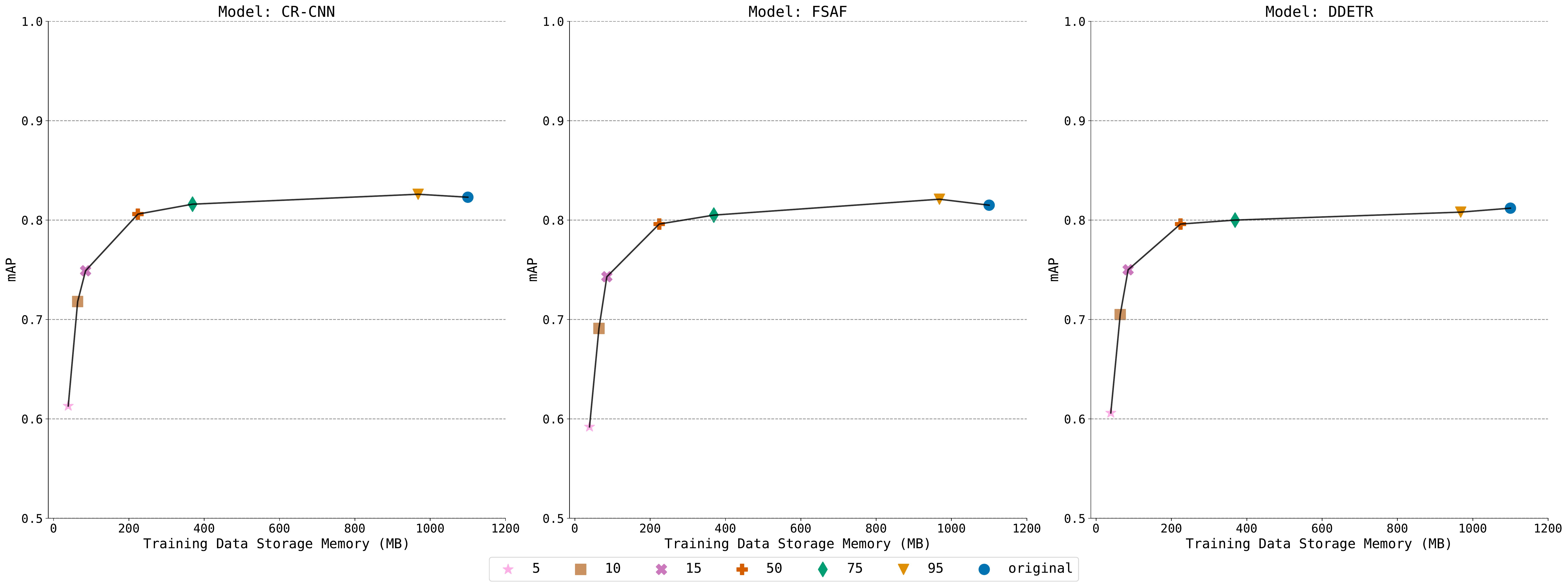}
    \vspace{-0.8cm}
    \caption{FLIR: CNN models performance (mAP) against training data memory storage (Megabyte) at different compression levels.}
    \label{fig:flir-mem_map}
\end{figure*}

\begin{figure*}[!htb]
    \centering
    \includegraphics[width=\linewidth]{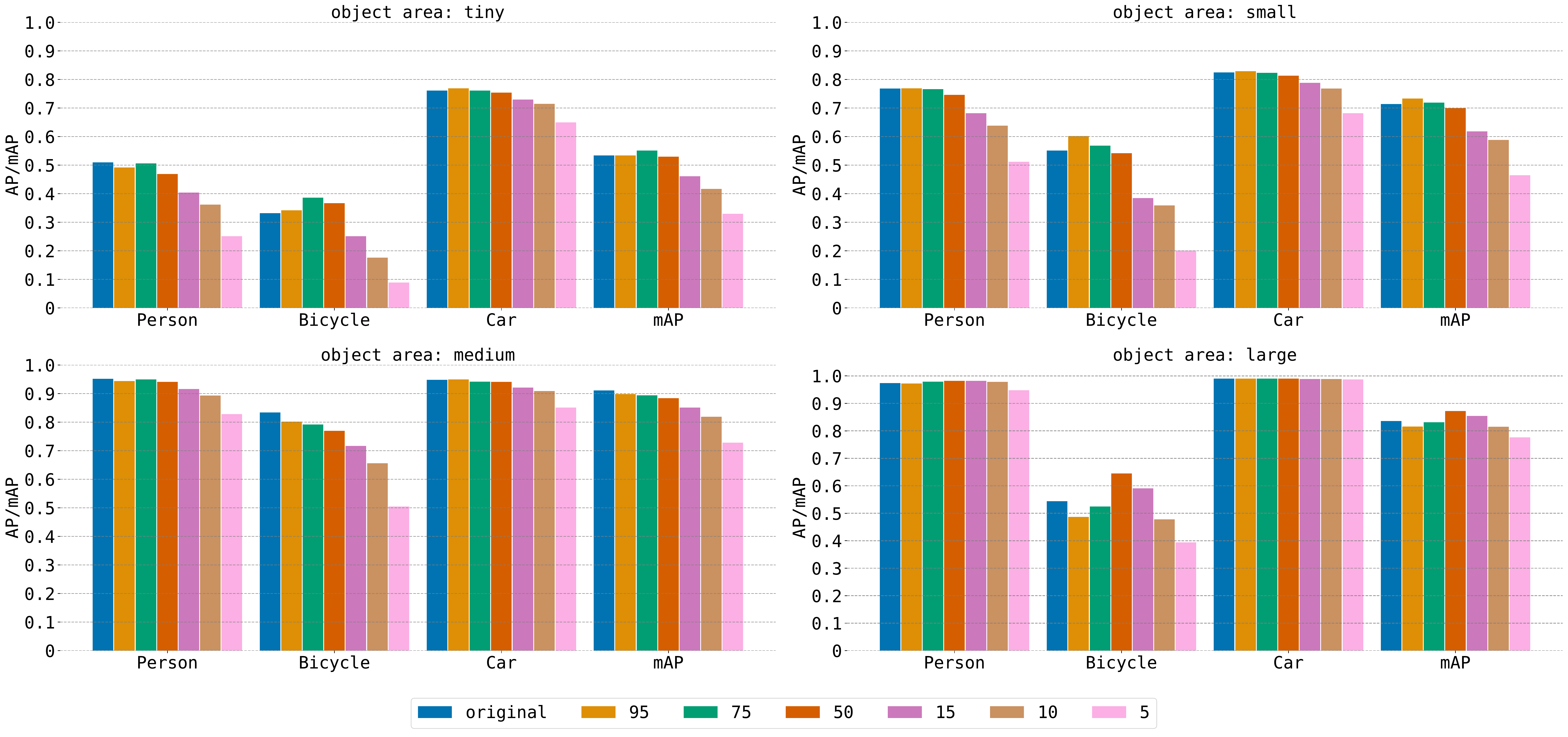}
    \vspace{-0.8cm}
    \caption{FLIR: Object area-wise AP/mAP analysis at different compression levels using Cascade R-CNN \cite{cai19:cascade}.}
    \label{fig:flir-crcnn}
\end{figure*}
\noindent
Table \ref{Table:lossy_flir} presents the results from evaluating CNN models (Section \ref{ss:objarch}) using {\it FLIR} dataset after undergoing JPEG lossy compression at six different quality levels: \{$95, 75, 50, 15, 10, 5$\}. As outlined in Section \ref{ss:expprotocol}, the models are first trained on the uncompressed/original dataset ({\it FLIR}) and evaluated on the compressed variants ({\it FLIR$\_c$}). Subsequently, the models are re-trained with the compressed imagery ({\it FLIR$\_c$}) at the respective level and evaluated on the compressed variants ({\it FLIR$\_c$}).

\begin{table}[!htb]
	\renewcommand*{\arraystretch}{0.90}
	\centering
	\vspace{-0.3cm}
	\caption{FLIR: Impact of lossy compression on object detection.}
	\vspace{-0.4cm}
	\addtolength{\tabcolsep}{-4.8pt}
	\begin{tabular}{|c|l|l|l|l|l|l|}
		\hline
		
		\multirow{3}{*}{\shortstack[c]{\scriptsize Compression \\ \scriptsize Level}} & \multicolumn{3}{c|}{{\it FLIR} $\Rightarrow$ {\it FLIR$\_c$}}  & \multicolumn{3}{c|}{{\it FLIR$\_c$} $\Rightarrow$ {\it FLIR$\_c$}} \\ \cline{2-7}
		&  \multicolumn{3}{c|}{mAP} &  \multicolumn{3}{c|}{mAP} \\ \cline{2-7}
		& \small CR-CNN & \small FSAF & \small DDETR & \small CR-CNN & \small FSAF & \small DDETR \\ \hline
		95 & 0.822 & 0.816 & 0.811 & 0.826 & 0.821  & 0.808 \\ 
		75 & 0.813 & 0.802 & 0.803 & 0.816 & 0.805  & 0.800 \\ 
		50 & 0.789 & 0.789  & 0.782 & 0.806 & 0.796 & 0.796 \\ 
		15 & 0.623 & 0.606 & 0.663 & 0.749 & 0.743 & 0.750 \\
		10 & 0.426 & 0.427 & 0.489 & 0.718 & 0.691 & 0.705  \\ 
		5 & 0.109 & 0.149 & 0.152 & 0.613 & 0.592 & 0.606 \\ \hline
		
	\end{tabular}
	\addtolength{\tabcolsep}{-0pt}  
	\label{Table:lossy_flir}
\end{table}

When the models are trained across the original dataset, and evaluated on compressed variants (Table \ref{Table:lossy_flir}, left), we observe that the models retain similar performance at the \{$95$\} compression level, only decreasing by $\sim1\%$ on average at the \{$75$\} compression level. At the compression level of \{$50$\}, the mAP is decreased only by $\sim3.5\%$ on average across all three models. We observe that a compression level of \{$15$\} and below has a greater impact on object detection performance. The model performances suffer significantly (mAP decreased by $\sim80\%$), when the images are compressed heavily, such as at \{$5$\} compression level across all three models. 

Re-training the models on lossy compressed imagery ({\it FLIR$\_c$}) at different compression levels and subsequently evaluating on the compressed variants ({\it FLIR$\_c$ $\Rightarrow$ FLIR$\_c$}) significantly ameliorated the performance (Table \ref{Table:lossy_flir}, right) for all three CNN models at higher compression levels. Re-training at the compression levels of \{$95, 75, 50$\} does not fully recover the performance compared to original training and evaluation protocol ({\it FLIR  $\Rightarrow$ FLIR$\_c$}). However, the impact of lossy compression at higher compression level (\{$5$\}), is mitigated through retraining with all models, achieving mAPs: \{$0.613, 0.592, 0.606$\} (Cascade R-CNN, FSAF, Deformable DETR correspondingly, Table \ref{Table:lossy_flir}, right), compared to uncompressed settings (mAP: \{$0.109, 0.149, 0.152$\}), an increment of $82\%$, $74\%$, and $74\%$ respectively. Amongst three models, two-stage Cascade R-CNN \cite{cai19:cascade}  offers superior performance across most of the compression levels, followed by transformer-based Deformable DETR \cite{zhu2021:deformable}, whilst one-stage FSAF \cite{zhu19:fsaf} are more robust to higher compression levels. Similar performance enhancement is discerned for compression levels of \{$15, 10$\}. The benefit of re-training on compressed images is depicted in Figure \ref{fig:flirdet_ex}, C, where compressed image trained models successfully detect the objects, contrary to the missing detection in Figure \ref{fig:flirdet_ex}, B.

Overall, the re-training improves the detection performance while affording a lossy JPEG compression rate much higher in terms of reduced image storage requirements, as illustrated in Figure \ref{fig:flir-mem_map}. We observe that the detection performance of the CNN models is resilient up to the compression level of \{$50$\}, while requiring only $1/5^{th}$ of training data storage overhead memory compared to uncompressed data storage ($224$ MB vs $1100$ MB, Figure \ref{fig:flir-mem_map}).  


\begin{figure*}[!htb]
    \centering
    \includegraphics[width=\linewidth]{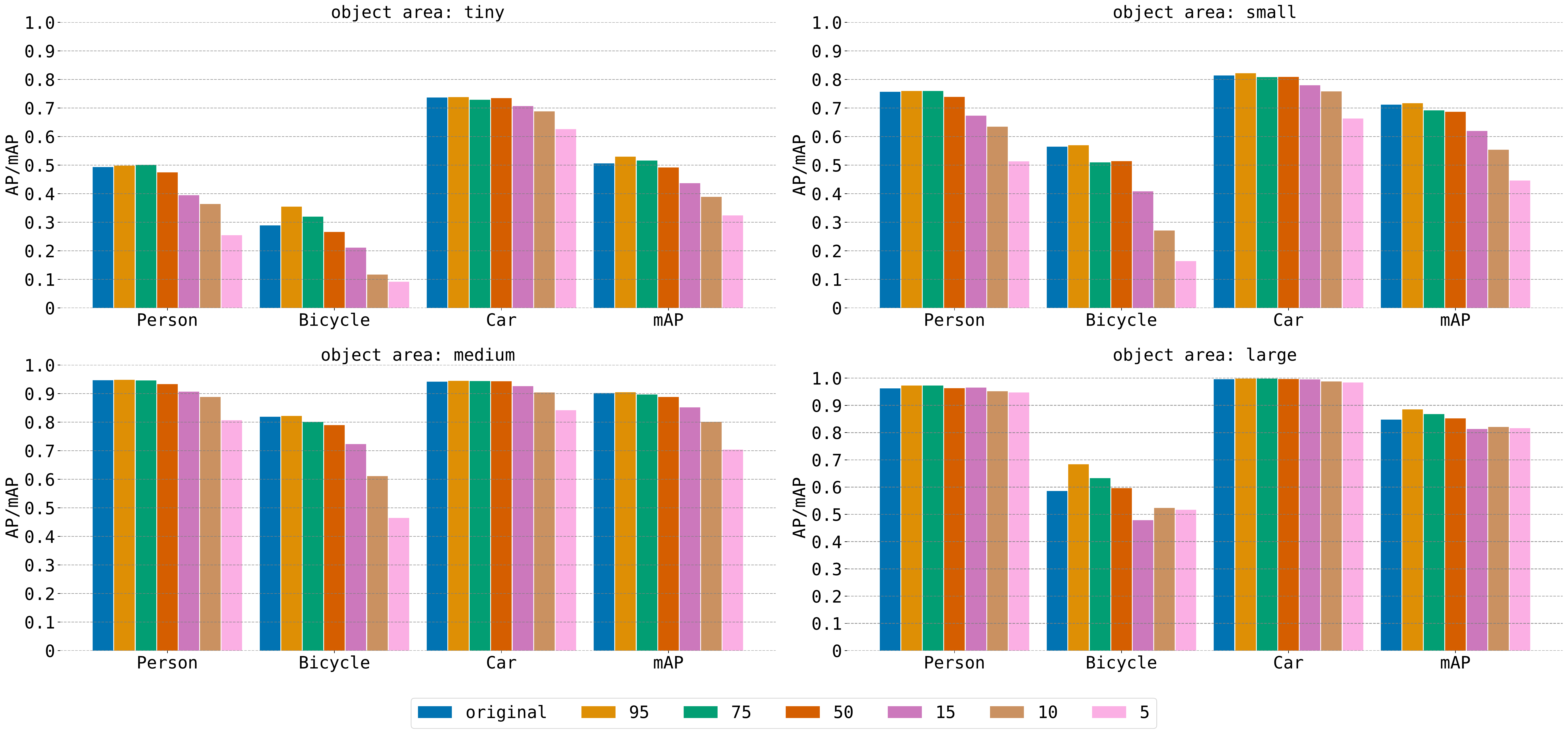}
    \vspace{-0.8cm}
    \caption{FLIR: Object area-wise AP/mAP analysis at different compression levels using FSAF \cite{zhu19:fsaf}.}
    \label{fig:flir-fsaf}
\end{figure*}

\begin{figure*}[!htb]
    \centering
    \includegraphics[width=\linewidth]{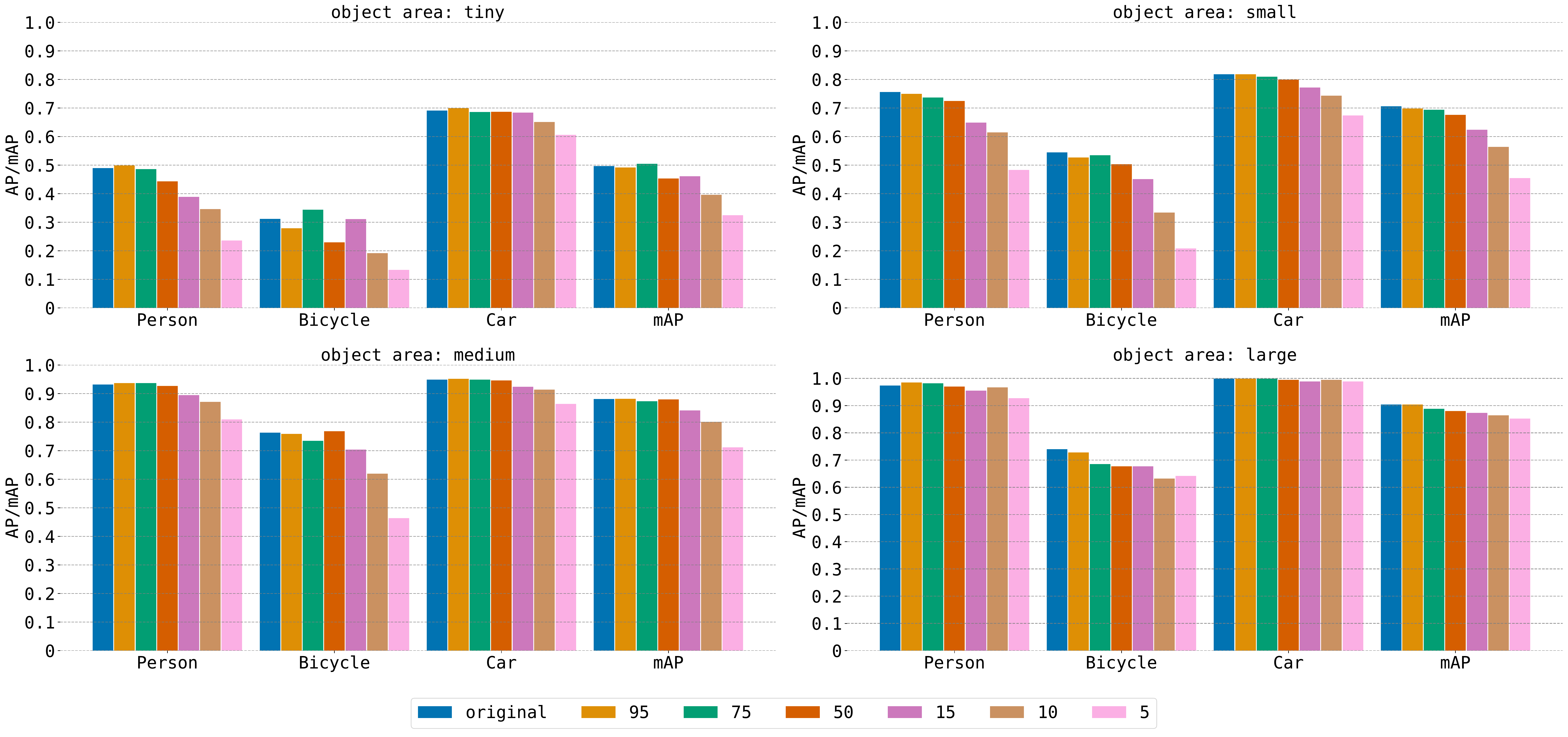}
    \vspace{-0.8cm}
    \caption{FLIR: Object area-wise AP/mAP analysis at different compression levels using Deformable DETR \cite{zhu2021:deformable}.}
    \label{fig:flir-ddetr}
\end{figure*}

\subsection{Object Area-wise Performance Analysis} \label{ss:objareamAP}
\vspace{-0.0cm}
\noindent
This section reports the in-depth analysis of the detection performance based on the discrete object area definitions (Section \ref{ss:objarea}) for each class in the dataset. First, we calculate AP for object area-wise using the uncompressed dataset setting ({\it FLIR $\Rightarrow$ FLIR}). Subsequently, we follow the compressed dataset setting ({\it FLIR$\_c$ $\Rightarrow$ FLIR$\_c$}) to obtain the object area-wise statistics as presented in Figures \ref{fig:flir-crcnn}, \ref{fig:flir-fsaf}, and \ref{fig:flir-ddetr}. 

The {\it FLIR} test-set comprises of $\sim 45\%$ {\it small} and $\sim 45\%$ {\it medium} area category objects (Table \ref{Tab:flirthermal_dbstat}) across three different classes. We observe that the impact of the heavy compression is more prevalent on {\it small} object areas than {\it medium}. The {\it small} object area mAP is reduced by $\sim 35\%$ on compressed imagery (at compression of $5$) compared to the uncompressed training settings across three CNN models (Figures \ref{fig:flir-crcnn}, \ref{fig:flir-fsaf}, and \ref{fig:flir-ddetr}), with FSAF \cite{zhu19:fsaf} suffering the most. On the other hand, the {\it medium} area performance is decreased by average $19\%$. The impact of lossy compression (at compression of $5$) is also noticeable on the {\it tiny} category with mAP reduced by $36\%$ on average. Nonetheless, the {\it large} category can withstand the maximal compression without severely affecting the detection performance. 

Amongst the three object classes of {\it FLIR} dataset, the {\it car} class is less impacted by the compression than {\it person, bicycle} across all object area categories. This is possibly due to {\it car} class having a higher percentage of {\it medium} and {\it large} area than other two classes. Bicycle was the worst performing class and the one most sensitive to compression. This is likely due to the reasons of: 1) The thin frames of bicycles become harder to see under higher degrees compression and 2) The rotational bias of bicycles (i.e. When a bike is viewed from the front rather than from the side, it appears thin and close to one-dimensional) and thus are harder to detect.  Furthermore, the performances of CNN models are less impacted at the lower compression levels (\{{$95, 75, 50$}\}) across all object area categories. Overall, Cascade R-CNN \cite{cai19:cascade} achieves the maximal mAP on {\it tiny, small, medium} object area category, whilst transformer-based Deformable DETR \cite{zhu2021:deformable} suited to {\it large} category for both uncompressed and compressed train and evaluation settings.









\vspace{-0.0cm}
\section{Conclusion} \label{sec:conclusion}
\vspace{-0.0cm}
\noindent
In this work we conduct an extensive study into the impact that lossy JPEG compression applied at differing discrete levels to infrared-band imagery has upon a common set of diverse object detection architectures. We evaluate the performance of three operationally diverse object detection methods, each with vastly differing detection approaches with respect to six discrete classes, \{{\it 95, 75, 50, 15, 10, 5}\}, of varying degrees of compression applied to the input data. Multi-stage Cascade R-CNN performs slightly better than single-stage and transformer-based architectures. However, single-stage FSAF, which has the smallest number of parameters, achieves the fastest inference time. Within this study, we report that significant compression between $75$ and $50$ will negligibly affect performance, but will reduce the storage capacity of the data by over $50$\%. We show that inferring models trained on uncompressed imagery will fail to detect objects across compressed images, but re-training the same models across compressed images will allow them to detect objects with high accuracy.
We also quantitatively report on how objects of differing size are affected by compression by assigning each object to a discrete class according to the bounding box area. We hope that the results presented in this work will help future real-world applications of object detection models applied to thermal imagery to reduce storage and transmission overhead while obtaining minimal performance impact.

\begingroup
\setstretch{.96}
{\small
\bibliographystyle{ieee_fullname}
\bibliography{egbib, object_detection}
}
\endgroup

\clearpage

\end{document}